\documentclass[sigconf]{acmart}
\AtBeginDocument{%
    }

\copyrightyear{2026}
\acmYear{2026}
\setcopyright{cc}
\setcctype{by}
\acmConference[MM '26]{Proceedings of the 34th ACM International Conference on Multimedia}{November 10--14, 2026}{Rio de Janeiro, Brazil}
\acmBooktitle{Proceedings of the 34th ACM International Conference on Multimedia (MM '26), November 10--14, 2026, Rio de Janeiro, Brazil}
\acmDOI{10.1145/3767308.3835532}
\acmISBN{979-8-4007-2213-4/2026/11}

\usepackage{subcaption}
\usepackage[ruled,vlined]{algorithm2e}
\usepackage{multirow}
\usepackage{colortbl}
\usepackage{adjustbox}
\usepackage{balance}

\definecolor{mygreen}{RGB}{0,0, 200}
\definecolor{myred}{RGB}{0, 200, 200}

\begin{document}

\title{TDVR: Joint Text Disambiguation and Viewpoint Reasoning for Zero-Shot 3D Visual Grounding}

\author{Qingxi Du}
\authornote{Both authors contributed equally to this research.}
\orcid{0009-0005-0172-9228}
\affiliation{%
  \institution{Northwestern Polytechnical University}
  \department{Software School}
  \city{Xi'an}
  \state{Shaanxi}
  \country{China}
}
\email{duqingxi@mail.nwpu.edu.cn}

\author{Junbo Wang}
\authornotemark[1]
\orcid{0009-0006-9955-7838}
\affiliation{%
  \institution{Northwestern Polytechnical University}
  \department{Software School}
  \city{Xi'an}
  \state{Shaanxi}
  \country{China}
}
\email{jbwang@nwpu.edu.cn}

\author{Yuke Li}
\authornote{Corresponding author.}
\orcid{0000-0001-9836-4600}
\affiliation{%
  \institution{Northwestern Polytechnical University}
  \department{Software School}
  \city{Xi'an}
  \state{Shaanxi}
  \country{China}
}
\email{liyuke@nwpu.edu.cn}

\author{Yining Zhu}
\authornotemark[2]
\orcid{0009-0006-6003-7739}
\affiliation{%
  \institution{Northwestern Polytechnical University}
  \department{School of Computer Science}
  \city{Xi'an}
  \state{Shaanxi}
  \country{China}
}
\email{yiningzhu@nwpu.edu.cn}

\renewcommand{\shortauthors}{Qingxi Du, Junbo Wang, Yuke Li, and Yining Zhu}

\begin{abstract}
Zero-shot 3D visual grounding aims to localize specific objects based on textual descriptions and 3D visual input, playing an important role in many fields such as embodied intelligence. However, the effectiveness of existing methods is significantly hindered by the ambiguous query text and deficient viewpoints.
To address these issues, we propose TDVR, a training-free reasoning framework that disambiguates the input text and infers accurate viewpoints for zero-shot 3D visual grounding. First, we construct semantic 3D scene graph from the detected instances in the 3D point cloud.Subsequently, we put the original query, appearance and spatial relationship descriptions into the LLM for fusion, thereby disambiguating the initial input.Given the complexity of linguistic expressions, we leverage chain-of-thought reasoning to generate the structured representation of disambiguated query.Then taking the scene graph and structured query as input, we get the optimal view via viewpoint reasoning to solve the problem of missing viewpoints during grounding.Based on the obtained optimal viewpoint,we further discriminate the distracting objects,enabling the model with the ability to distinguish similar instances.After that, we match the category text and appearance images with the query by computing the similarity of feature vectors.Finally  the target object was identified by integrating the viewpoint score,confusion score,category score,and appearance score.Compared with previous methods, our TDVR has stronger capabilities in viewpoint reasoning, similar object discrimination, and ambiguous query understanding.Experimental results on the public ScanRefer dataset show that our method outperforms the existing state-of-the-art methods by 15.25\% and 14.46\% in Acc@0.25 and Acc@0.5 respectively, demonstrating the effectiveness of our TDVR in addressing ambiguous query text and deficient viewpoints. 
\end{abstract}




\maketitle
\section{Introduction}
3D Visual Grounding (3DVG) is a fundamental task in Embodied AI \cite{liu2025reasongrounder,lu2024scaneru} and autonomous driving\cite{li2025nugrounding,wang2025omnidrive}, requiring agents to precisely localize objects within complex 3D environments based on natural language. Unlike 2D methods \cite{lin2023uninext,rasheed2024glamm,ma2024groma,man2025argus,chen2023shikra}, 3DVG involves intricate geometric topologies and viewpoint-dependent spatial relations, posing significant challenges for cross-modal alignment and geometric reasoning.

\begin{figure}[t]
    \centering
    \begin{minipage}{0.85\columnwidth} 
        \centering
        
        \begin{subfigure}[b]{0.25\linewidth}
            \centering
            \includegraphics[width=\linewidth]{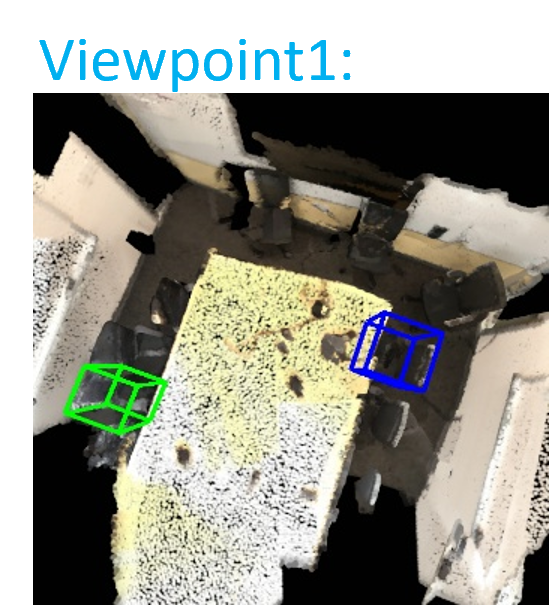}
        \end{subfigure}
        \begin{subfigure}[b]{0.25\linewidth}
            \centering
            \includegraphics[width=\linewidth]{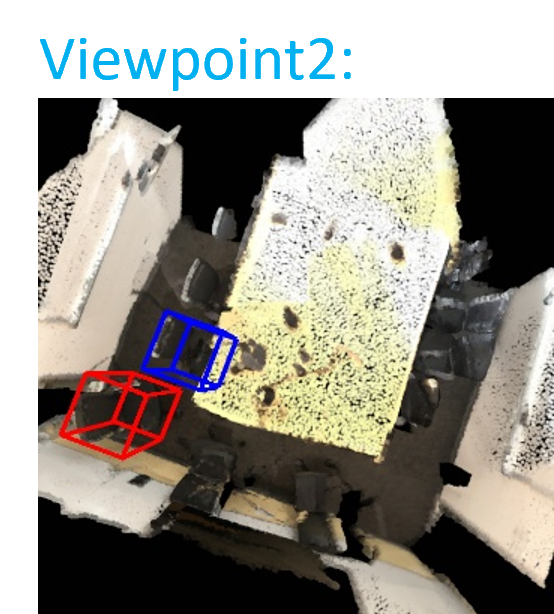}
        \end{subfigure}
        \hspace{10pt}
        \begin{subfigure}[b]{0.25\linewidth}
            \centering
            \includegraphics[width=\linewidth, height=1.7cm]{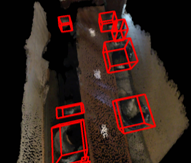}
        \end{subfigure}
        
        \vspace{2pt} 
        \small 
        \makebox[0.6\linewidth][c]{{Viewpoint-aware:\color{red}False}} 
        \hspace{10pt} 
        \makebox[0.28\linewidth][c]{{\color{red}Similarity Confusion}}\\ [4pt]
        
        \normalsize (a) Previous Method
        
        \vspace{2pt} 

        \begin{subfigure}[b]{0.25\linewidth}
            \centering
            \includegraphics[width=\linewidth]{problems_we_face/v1_t.png}
        \end{subfigure}
        \begin{subfigure}[b]{0.25\linewidth}
            \centering
            \includegraphics[width=\linewidth]{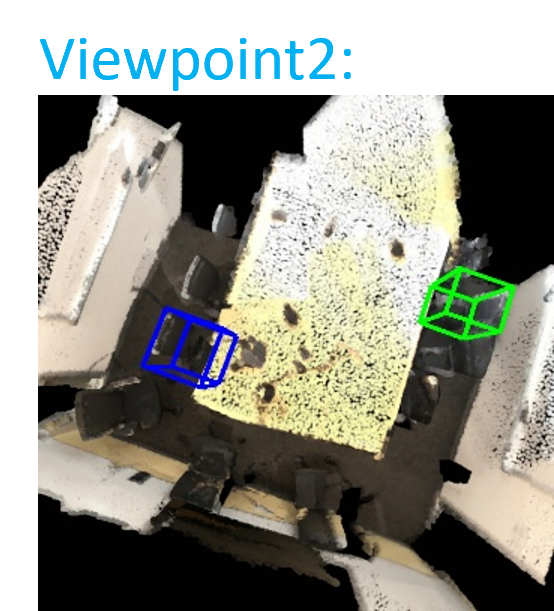}
        \end{subfigure}
        \hspace{10pt}
        \begin{subfigure}[b]{0.25\linewidth}
            \centering
            \includegraphics[width=\linewidth, height=1.7cm]{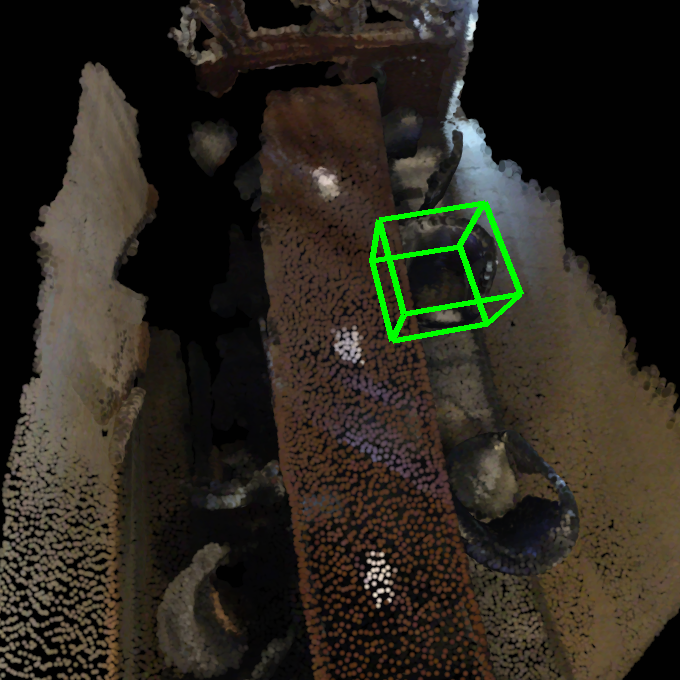}
        \end{subfigure}
        
        \vspace{2pt}
        \small
        \makebox[0.6\linewidth][c]{{Viewpoint-aware:\color{green!60!black}True}} 
        \hspace{10pt} 
        \makebox[0.28\linewidth][c]{{\color{green!60!black}Clear Discrimination}}\\ [4pt]
        \normalsize
        \normalsize (b) Our Method
    \end{minipage}

    \vspace{-10pt} 
    \caption{Under two challenging situations, performance comparison of different methods.(a) Previous methods yield incorrect localization.(b) Our method achieves correct localization.}
    \label{prblems}
    \Description{This image presents a qualitative comparison of 3D visual grounding results between a baseline approach and the proposed method on the ScanRefer dataset. The figure is organized into two horizontal rows, labeled (a) and (b), representing "Previous Method" and "Our Method" respectively.

Top Row (a) - Previous Method: This row displays failure cases where the model struggles with spatial reasoning. On the left, the visualization shows "incorrect localization," where the predicted bounding boxes (indicated in red or green) fail to align with the target object. On the right, the "random guess" scenario illustrates the model producing multiple overlapping red boxes across the scene, indicating high uncertainty and a lack of precise object identification.

Bottom Row (b) - Our Method: This row demonstrates the superior performance of the proposed framework. In the same scenes shown above, the model achieves "correct localization" and "accurate prediction," marked by precise green bounding boxes that closely fit the target objects within the complex 3D point cloud environment.

The comparison highlights the proposed method's ability to disambiguate target objects more effectively, leading to significantly more robust and accurate spatial grounding results compared to existing techniques.}
\end{figure}

Recently, 3D visual grounding methods \cite{feng2024naturally,wang2025augrefer,wang2025liba,guo2025text,xu2024vlm,zhang2024towards,yuan2024visual,li2025seeground,lin2025seqvlm,mi2025language,jin2025spazer} have achieved remarkable performance, yet they are still facing two major challenges:
\textbf{(1)Spatial ambiguity with unknown viewpoints.} 
Spatial relationship descriptions are often view-dependent, such as "on the left" and "at the back". 
However, in 3D space, these directional terms will lead to severe referential ambiguity if the describer’s camera pose is not explicitly specified.
As shown in the two images on the left side of Fig.\ref{prblems}a,
if we localize according to the query “the office chair left of the armchair (marked by the blue box)”, the localization may be accurate under viewpoint 1.However, when the viewpoint is rotated 180° to viewpoint 2, the description “on the left” will result in an incorrect target localization.
Most existing models fail to account for the influence of viewpoint.In contrast, as shown in Fig. \ref{prblems}b, benefiting from viewpoint reasoning, our model achieves accurate object localization across different views.
\textbf{(2)Ambiguous queries under similar instance interference.} 
In dense indoor scenes, there often exist multiple distractors with the same category, similar appearance, and close proximity.As shown in  the right image of Fig. \ref{prblems}a, given the query "A black leather armchair next to another identical chair", there exist multiple objects that match the description.
Therefore, when the textual description is relatively brief and multiple similar objects exist in the scene, most agents lack fine-grained discrimination method.
They can only rely on random guessing among multiple candidates, making it difficult to achieve robust and deterministic localization.In contrast, as shown in Fig.\ref{prblems}b, our model can accurately localize objects in scenes with multiple object interference.

To address these issues, we propose TDVR, a training-free reasoning framework that disambiguates the input text and infers accurate viewpoints for zero-shot 3D visual grounding. 
First, from the detected instances and the 2D images cropped according to minimum occlusion, we construct the semantic 3D scene graph.
Meanwhile, we feed the original query, the appearance descriptions generated from 2D images, as well as the spatial relationship descriptions into the LLM for fusion. By doing so, we disambiguate the original query.
For these disambiguated queries, we perform structured extraction via chain-of-thought reasoning and result self-checking,ensuring they accurately capture the semantics of the description.
After that, taking the scene graph and structured query as input, we infer the viewpoints that best match the description by rotating the point cloud and simultaneously compute the viewpoint score for each object.
Based on the obtained optimal viewpoint, we calculate the confusion score of each object by evaluating how its spatial relationship with other intra-class instances matches the description, enabling the model to distinguish similar instances.
Subsequently, we match the category text and appearance images with the query by computing the similarity of feature vectors.
Finally  the target object is identified by integrating the viewpoint score,confusion score,category score,and appearance score.

We conducted experiments on the Sr3D \cite{achlioptas2020referit3d} and ScanRefer \cite{chen2020scanrefer} datasets.
On the Sr3D\cite{achlioptas2020referit3d} dataset, we achieve an overall accuracy of 70\%, and 79.03\% accuracy on the viewpoint-dependent subset.
Experimental results on the ScanRefer \cite{chen2020scanrefer} dataset demonstrate that TDVR sets a new state-of-the-art (SOTA) for zero-shot methods, reaching 64.06\% on the strict Acc@0.5 metric—a 14.46\% improvement over the previous best. Remarkably, TDVR even outperforms several recent fully supervised models (e.g., TSP3D \cite{guo2025text}, Pseudo-EV \cite{geng2025pseudo}), significantly narrowing the gap between zero-shot reasoning and supervised learning.

Our primary contributions are summarized as follows:
\begin{itemize}
\item We propose TDVR, a zero-shot 3D visual grounding framework based on MLLM for ambiguity elimination and viewpoint inference.
\item We introduce a query disambiguation pipeline,which enhances the descriptive text by integrating expressions such as appearance and spatial relationship.
\item We propose a method for viewpoint inference and similar object discrimination, which enhances the model's spatial perception ability in complex scenes with multiple distractors.
\item We achieve substantial performance improvements on ScanRefer and Sr3D,demonstrating excellent localization capability.
\end{itemize}

\section{Related Work}
\subsection{Supervised 3D Visual Grounding }
Many supervised 3DVG methods have also achieved favorable performance. Some are combined with LLMs, some incorporate viewpoint information, and others employ voxel cropping, all of which have yielded promising results.
Pseudo-EV \cite{geng2025pseudo} resolves viewpoint ambiguity by splitting 3DVG into embodied viewpoint prediction and target localization. It adopts LLM-generated pseudo-labels and semantic structuring to cut computation costs and boost localization accuracy.
DDPA-3DVG \cite{gu2025ddpa} achieves coarse-to-fine progressive cross-modal alignment by performing dual disentanglement and multi-grained feature extraction on both 3D scenes and natural language.
TSP3D \cite{guo2025text} is an efficient single-stage 3D visual grounding framework. Based on multi-level sparse convolution, it employs text-guided pruning to refine voxels and performs compensation to repair over-pruning, achieving both superior accuracy and real-time speed.
VPP‑Net  \cite{shi2024aware}  explicitly predicts the speaker's viewpoint and rotates the 3D scene, assisted by a unified object representation loss, achieving state-of-the-art  performance across multiple datasets.

\subsection{Zero-shot 3D Visual Grounding}
Existing zero-shot 3DVG methods adopt LLMs and MLLMs. Utilizing their cross-modal vision-language ability, these approaches convert 3D point clouds to 2D renders or scene video frames, and input such visuals into multimodal models to build vision-text alignment.
ZSVG3D \cite{yuan2024visual}  propose a zero-shot 3D visual grounding method based on visual programming, which uses LLMs to generate code and combines 2D/3D modules to achieve complex spatial reasoning. 
SeeGround \cite{li2025seeground}'s core idea is to convert 3D scenes into a hybrid representation of query-aware 2D renderings and 3D spatial textual descriptions, and strengthen the alignment between visual and spatial information through visual prompting techniques.
SeqVLM \cite{lin2025seqvlm} leverages visual-language models to perform spatial reasoning and feature fusion guided by multi-view image sequences and 2D candidate boxes, thereby achieving high-precision zero-shot 3D visual grounding.
LASP \cite{mi2025language} employs an LLM to translate natural language instructions into executable 3D spatial programming codes, and directly parses target objects and their complex spatial relations in 3D space using mathematical logic, realizing efficient and high-precision training-free 3D visual grounding.
SPAZER \cite{jin2025spazer} achieves zero-shot 3D visual grounding through progressive reasoning by combining 3D spatial analysis and 2D semantic verification. It uses a VLM agent to collaborate on multimodal information, improving the robustness of localization without requiring 3D annotated data.

\section{Methodology}
\begin{figure*}[t]
  \centering
  \includegraphics[width=0.8\textwidth]{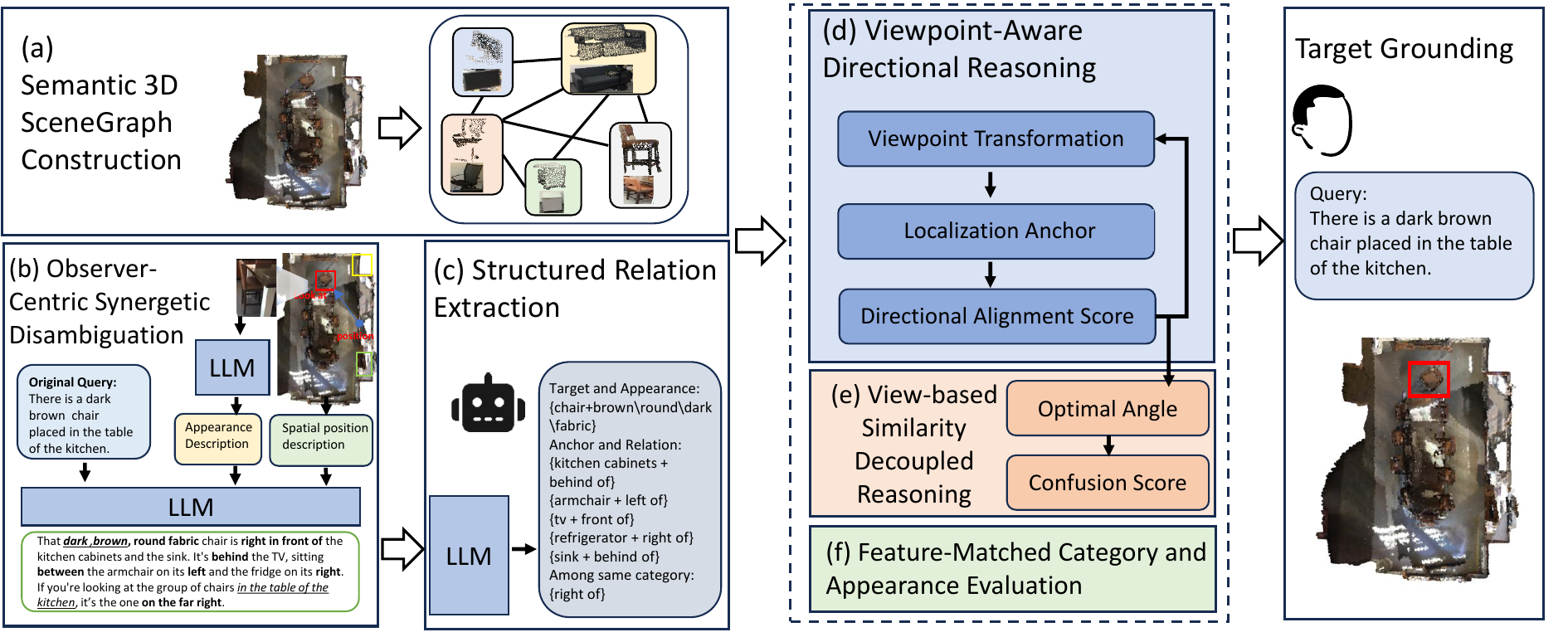}  
  \Description{This figure shows the flowchart of TDVR (Disambiguated Multi-dimensional Scene Graph Reasoning), a 3D visual grounding framework. Starting from the input 3D point cloud and the original query, the framework achieves precise target localization through multi-dimensional disambiguation and structured reasoning.
A detailed description of the flowchart structure is as follows:
Input Phase
3D point cloud: The raw 3D point cloud data of the input scene, showing an indoor space containing various furniture.
Multi-Dimensional Linguistic Disambiguation: As a core module of the framework, it uses a Large Language Model (LLM) to rewrite the vague original query into a richer and more detailed description. By integrating appearance descriptions and spatial position descriptions, the LLM generates an elaborate text covering target attributes, anchor objects, and their complex spatial relationships.
Structured Processing Phase
Scene Graph Construction: Converts the 3D scene into a scene graph, where nodes represent object instances and edges represent relationships between objects, modeling the topological structure of the environment.
Structured Relation Extraction: Extracts structured information from the enhanced text generated by the LLM, including:
Target and appearance: such as color, material, and shape (e.g., chair + brown/round/dark/fabric).
Anchors and relations: determines the target position relative to other furniture such as cabinets, TVs, and refrigerators.
Intra-class discrimination: further clarifies viewpoints among objects of the same category.
Reasoning & Discrimination Phase
This stage consists of three parallel logical modules:
viewpoint-Aware Directional Reasoning
Intra-class Confusion Discrimination, which resolves ambiguities caused by multiple similar objects in the scene.
Category and Appearance Matching
Output Phase
Grounding Target: Outputs the final localization result. In the 3D point cloud visualization, the target object is highlighted with a distinct red 3D bounding box.
}
  \caption{Overview of TDVR. TDVR begins by refining the raw query via the Observer-Centric Synergetic Disambiguation module, while constructing a scene graph to organize the 3D point cloud. The Viewpoint-Aware Directional Reasoning module then predicts the optimal viewpoint and score. Subsequently, the framework evaluates targets through View-based Similarity Decoupled Reasoning, Feature-Matched Category and Appearance Evaluation modules. By integrating these scores through weighted summation, the system identifies the object with the highest overall confidence as the final target.}
  \label{fig:pipeline}
\end{figure*}
\subsection{Overview}
As shown in Fig. \ref{fig:pipeline},we first disambiguate the original query via the proposed observer-centric synergetic disambiguation.
The disambiguated query is fed into the structured relation extraction module to extract textual semantics.
Then ,we construct a scene graph $G$ for objects in the scene to represent their multiple attributes and spatial relationships.
After that, the scene graph is fed into the viewpoint-aware directional reasoning module to obtain the optimal viewpoint score $S_v$ and the inferred optimal viewpoint list $\Theta_{best}$ . Based on the inferred view, we rotate the point cloud and calculate the confusion score $S_c$  for each object via the view-based similarity decoupled reasoning module. Subsequently, we compute the category matching score $S_{cat}$ and appearance matching score $S_{app}$ via the feature-matched category and appearance evaluation module.
Finally, we calculate an overall score by weighted summation across multiple metrics, then select the object with the highest score as the predicted target.

\subsection{Semantic 3D Scene Graph Construction}

We represent the 3D scene as a scene graph $G = (V, E, \mathcal{X}_V)$ \cite{rosinol20203d,wu2021scenegraphfusion,gu2024conceptgraphs,chang2023context,zemskova20253dgraphllm,yin2024sg}, where each node $v_i \in V$ corresponds to a detected instance. The node attributes are defined as $\mathbf{x}_i = [p_i, a_i, l_i]$, where $p_i \in \mathbb{R}^3$ denotes the 3D bounding box, $l_i$ is the category label, and $a_i$ is an optimally selected 2D cropped view.

We employ a pre-trained detector to extract object candidates and their spatial geometric properties. For each candidate, we generate the 2D attribute $a_i$ by identifying the viewpoint with minimal occlusion. Specifically, we project the 3D point cloud $\mathbf{P}_{world}$ onto the image plane $(u, v)$ using the camera's intrinsic $K$ and extrinsic $T_{w2c}$ matrices:

$$\mathbf{P}_{pixel} = K \cdot (T_{w2c} \cdot \mathbf{P}_{world})$$

To ensure $a_i$ provides high-quality visual features, we implement a depth-consistency check. A point is considered visible if its calculated depth $d_{calc}$ aligns with the recorded depth map value $d_{map}$ within an extremely small threshold $\epsilon$: $|d_{calc} - d_{map}| < \epsilon$. We select the 2D crop from the frame that yields the highest ratio of visible points for the target object. As shown in Algorithm \ref{alg:scene_graph}, we present the pseudocode for our semantic 3D scene graph construction algorithm.

\begin{algorithm}[t]
\footnotesize
\caption{Semantic 3D Scene Graph Construction}
\label{alg:scene_graph}
\KwIn{Point cloud $\mathbf{P}_w$, RGB image $\mathcal{F}$,depth map $\mathcal{D}_f$ , Matrices $\{K, T_{w2c}\}$, Threshold $\epsilon$.}
\KwOut{Scene Graph $G = (V, E, \mathcal{X}_V)$.}

Detect object candidates $\mathcal{O} = \{o_1, \dots, o_n\}$\;
\ForEach{$o_i \in \mathcal{O}$}{
    $V_{max} \leftarrow 0$; \quad $p_i, l_i \leftarrow \text{Extract3D}(o_i)$\;
    \ForEach{frame $f \in \mathcal{F}$}{
        $\mathbf{P}_{px} = K \cdot (T_{w2c} \cdot \mathbf{P}_w)$\;
        $N_{vis} = \sum_{\mathbf{p} \in \mathbf{P}_w} \mathbb{I}(|z(\mathbf{p}_{cam}) - \mathcal{D}_f(u, v)| < \epsilon)$\;
        \If{$Ratio = N_{vis} / |\mathbf{P}_w| > V_{max}$}{
            $V_{max} = Ratio$; \quad $a_i = \text{Crop}(f, o_i)$\;
        }
    }
    $v_i \leftarrow [p_i, a_i, l_i] \in V$\;
}
Construct $E$ via spatial proximity\;
\Return $G = (V, E, \mathcal{X}_V)$\;
\end{algorithm}

\subsection{Observer-Centric Synergetic Disambiguation}

To resolve the inherent ambiguities in human-provided descriptions, we propose the observer-centric synergetic disambiguation module. This module enhances the original query by generating three types of descriptions: appearance details of the target object, directional descriptions of multiple anchors, and spatial cues for instances of the same category.

First, we extract fine-grained appearance descriptions (e.g., color, texture) by feeding 2D object crops into an MLLM. Next, to determine the spatial relations from the observer's perspective, we establish a local coordinate system. Let $V_{gt} = L - P$ be the viewing direction from camera position $P$ to look-at point $L$. We project $V_{gt}$ onto the floor plane to define the horizontal heading $V_{hgt}$ as the Y-axis, with $P$ as the origin. To optimize computation, we transform only the 3D bounding boxes into this observer-centric space rather than the entire point cloud. We resolve horizontal relative directions by selecting $n$ anchor points across four sectors:

$$A_{dir} = \{ (dx, dy) \mid |d_{axis1}| \ge |d_{axis2}|, \, d_{axis1} \gtrless 0 \}$$

For each anchor, we compute the cosine similarity between the target-to-anchor vector $v_{tar2an}$ and eight canonical directional vectors . The direction with the maximum similarity is embedded into a template: "[anchor] is [direction] of [target]".
To distinguish the target from distractors of the same class, we generate intra-category descriptions.
We construct a category-specific local frame centered at the mean coordinates of all same-category objects and determine the target’s relative position. Subsequently, according to the obtained relative position, we generate position descriptions among instances of the same category (e.g., "Among these [target objects], the one I want is the one located at the [direction]").
Note that vertical relationships are determined by direct Z-axis comparison ,so no additional enhancement is required.
Finally, the LLM integrates the original query with these enriched descriptions to produce a precise, disambiguated query for grounding.

\subsection{Structured Relation Extraction}
To bridge the gap between enriched natural language and machine-interpretable reasoning, we employ an LLM as a Structured Information Extractor. 
We propose a Chain-of-Thought  prompting strategy \cite{wei2022chain,kojima2022large,wang2022self,lu2023survey,zhao2023survey,ho2023large} that systematically decomposes the query into a structured semantic representation through the following four inference stages: 
\textbf{Stage 1}: Target Entity Identification. The LLM identifies the primary target subject $T$ and extracts its fine-grained appearance attributes description $\mathcal{A}_T$ (e.g., color, material, and texture).
\textbf{Stage 2}: Anchor Object Localization. The model parses the description to identify reference objects or "anchors" $S_{anc}$, which serve as spatial landmarks to constrain the search space.
\textbf{Stage 3}: Spatial Relation Standardization. Descriptive spatial cues are normalized into a set of predefined binary tuples $\mathcal{R} = \{ \langle s_i, r_i \rangle \}$, where $s_i \in S_{anc}$ and $r_i$ represents a standardized directional or proximity relation (e.g., left-of, under).
\textbf{Stage 4}: Intra-category Refinement. To resolve ambiguities among distractors of the same class, the LLM extracts the target's relative viewpoint within its category.
To ensure the integrity of the output, the large language model is required to re-verify the parsed structured expressions against predefined formatting specifications. If the output fails to meet the format requirements, the extraction is performed again. This ensures that the extracted expressions are maximally accurate and provides a deterministic semantic foundation for subsequent 3D visual grounding.

\subsection{Viewpoint-Aware Directional Reasoning}
\begin{figure}[htbp]
  \centering
  \includegraphics[width=\linewidth]{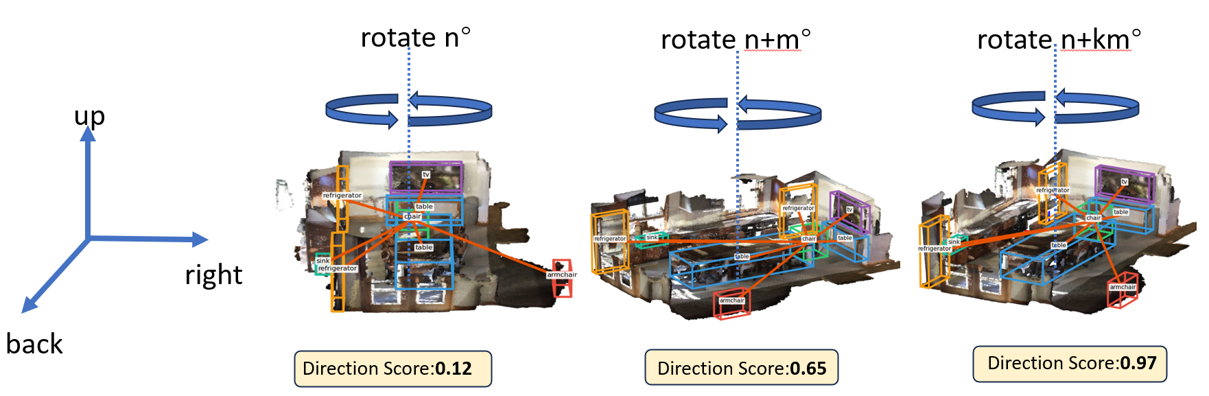}
  \caption{Schematic diagram of the Viewpoint-Aware Directional Reasoning module:
The left side shows the coordinate system for direction judgment, and the right side presents the view scores of the target object under different viewpoints.}
  \label{fig:OADR}
  \Description{A 3D coordinate system is defined on the left side of the figure, including three axes: up, right, and back. The main plot on the right is divided into three sequences, showing the viewpoint recognition of the same 3D reconstructed indoor scene under different rotation angles.
Viewpoint Variation and Rotation
The scene is rotated by n, n+m, and n+km respectively. As the rotation angle changes, the model’s viewpoint of the objects shifts significantly.
Object Relationship Modeling
Multiple colored 3D candidate bounding boxes are annotated in each scene graph (e.g., orange for refrigerator, red for chairs, etc.). Orange connection lines are emitted from the central target object as the origin to surrounding anchor objects, demonstrating that the model is attempting to construct a graph of relative spatial relationships between objects.
Direction Score
A score is attached below each sub-figure to quantify the matching degree between the textual description and the actual spatial viewpoint under the current viewpoint. At a rotation angle of n
, the score is 0.12, indicating low alignment in the current viewpoint. As the angle rotates, the score gradually increases to 0.65. When rotated to n+km
, the score reaches its maximum of 0.97, meaning the relative viewpoint of the objects under this viewpoint are highly consistent with the text description.}
\end{figure}

This section proves the rotation invariance of object spatial relationships to support viewpoint-aware directional reasoning. Specifically, the relative spatial relationship between objects is position-irrelevant under a fixed rotation angle, enabling viewpoint estimation via point cloud rotation.

Given an arbitrary 3D point $\mathbf{p} \in \mathbb{R}^3$, the point rotated around center $\mathbf{c}$ with rotation matrix $R \in \mathrm{SO}(3)$ is formulated as:
\begin{equation}
    \mathbf{p}' = R(\mathbf{p} - \mathbf{c}) + \mathbf{c}
\end{equation}
Let $\mathbf{p}_A$ and $\mathbf{p}_B$ denote the centroids of object $A$ and $B$, respectively. The rotated centroids are:
\begin{equation}
    \mathbf{p}_A' = R(\mathbf{p}_A - \mathbf{c}) + \mathbf{c}, \quad \mathbf{p}_B' = R(\mathbf{p}_B - \mathbf{c}) + \mathbf{c}
\end{equation}
The relative displacement vector after rotation is derived as:
\begin{equation}
    \mathbf{v}' = \mathbf{p}_B' - \mathbf{p}_A' = R(\mathbf{p}_B - \mathbf{p}_A) = R\mathbf{v}
\end{equation}
The derivation indicates that the transformed relative vector $\mathbf{v}'$ is only determined by the rotation matrix $R$ and the original relative vector $\mathbf{v}$, which is independent of the rotation center $\mathbf{c}$. Accordingly, with a fixed rotation matrix $R$ and direction judgment vector $\mathbf{u}$, the projection score characterizing the relative spatial relationship remains stable:
\begin{equation}
    S = \langle \mathbf{v}', \mathbf{u} \rangle = \langle R\mathbf{v}, \mathbf{u} \rangle
\end{equation}
Therefore, viewpoint inference is independent of object position. Combined with the vertical direction prior in Section 3.3 that $Z$-axis-based vertical judgment is unambiguous without pitch optimization, the original 6-D viewpoint inference can be reduced to 1-D horizontal rotation angle estimation.

The core of viewpoint-aware directional reasoning is to retrieve the optimal rotation viewpoint that matches the textual spatial constraint. We randomly sample $\mathbf{p}_{\mathrm{rand}}$ as the rotation center and rotate the entire point cloud with horizontal angle $\theta$ to simulate diverse observation perspectives:
\begin{equation}
    \mathbf{p}'_k(\theta) = R_z(\theta) \cdot (\mathbf{p}_k - \mathbf{p}_{\mathrm{rand}})
\end{equation}
The rotated relative vector between the target object $T$ and anchor object $A_i$ is:
\begin{equation}
    \vec{v}_{TA_i}(\theta) = \mathbf{p}'_{A_i}(\theta) - \mathbf{p}'_T(\theta)
\end{equation}
For each spatial constraint in structured queries, we first match anchor objects of category $C_j$ via category feature similarity and obtain the standard reference vector $\vec{v}_{\mathrm{ref},j}$ corresponding to the spatial relation. The alignment degree between the rotated relative vector and the reference vector is defined as:
\begin{equation}
    \Psi(T, C_j, \theta) = \max_{A_i \in C_j} \left\{ \operatorname{ReLU} \left( \frac{\vec{v}_{TA_i}(\theta) \cdot \vec{v}_{\mathrm{ref},j}}{\|\vec{v}_{TA_i}(\theta)\|} \right) \right\}
\end{equation}
The $\operatorname{ReLU}$ function eliminates negative matching interference, and max pooling selects the best-matched anchor from similar distractors. To avoid premature filtering of valid targets, we calculate viewpoint scores for all candidate objects instead of screening high-score objects in advance. The final optimal viewpoint score is obtained by traversing all candidate rotation angles:
\begin{equation}
    S_v = \max_{\theta \in \Theta} \sum_{j=1}^{N} \Psi(T, C_j, \theta)
\end{equation}
This paradigm efficiently generates the optimal viewpoint set $\Theta_{\mathrm{best}}$ for each target object while calculating the viewpoint matching score $S_v$.

\subsection{View-based Similarity Decoupled Reasoning}

\begin{figure}[htbp]
  \centering
  \includegraphics[width=0.65\linewidth]{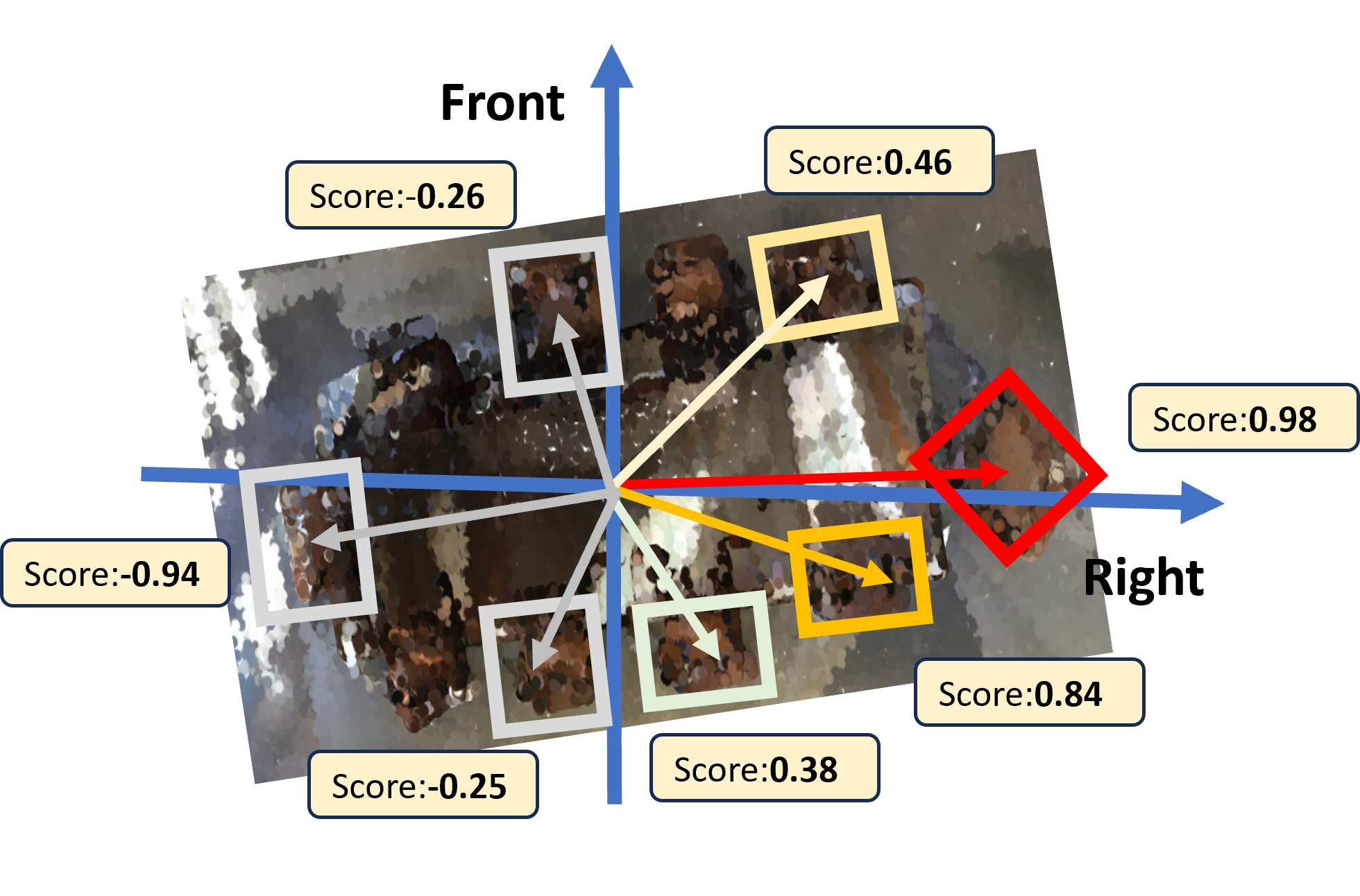}
  \caption{Schematic diagram of the View-based Similarity Decoupled Reasoning module.This figure illustrates the confusion scores of different chairs in the scene.}
  \label{fig:ICCR}
  \Description{This figure illustrates the viewpoint-aware target matching mechanism in 3D visual grounding. The background is a top-down 3D point cloud scene overlaid with a blue coordinate system centered at an origin point, where the vertical axis is labeled Front and the horizontal axis Right.
Multiple colored 3D candidate bounding boxes are distributed across different quadrants of the coordinate system, each connected to the central origin by an arrow of a corresponding color. A score indicating the matching confidence is labeled next to each box. As can be observed, the red bounding box located exactly to the positive right along the axis achieves the highest score of 0.98, while a candidate box to its left-rear obtains the lowest score of only −0.94.
This score distribution intuitively demonstrates how the model screens out the object that best matches the spatial description “directly to the right” from numerous candidates by computing the alignment of spatial vectors, thereby realizing high-precision viewpoint discrimination and object disambiguation.}
\end{figure}

While viewpoint reasoning (Section 3.5) evaluates candidates based on global spatial relations, multiple adjacent objects of the same category often satisfy the same directional constraints, leading to intra-class confusion. To distinguish the target from these similar distractors, as shown in Fig. \ref{fig:ICCR}, we introduce a fine-grained reasoning module centered on the relative distribution of the category.

For a given category $C$, let $\mathcal{S}_C = \{O_1, O_2, \dots, O_n\}$ be the set of all instances of the same category as the detected target object in the scene. To establish a stable internal reference, we compute the category's geometric centroid $\mathbf{M}_C$ in the rotated coordinate system.The rotation angle here is derived from the optimal view obtained for each object in Section 3.5.
$$\mathbf{M}_C = \frac{1}{n} \sum_{O_j \in \mathcal{S}_C} \mathbf{p}'_j$$
where $\mathbf{p}'_j$ represents the rotated center coordinates of object $O_j$. For each candidate $O_i \in \mathcal{S}_C$, we define a local displacement vector $\vec{v}_{conf}$ relative to this class centroid:$$\vec{v}_{conf} = \mathbf{p}'_i - \mathbf{M}_C$$

Leveraging the intra-category directional cues parsed in Section 3.4(e.g., "the one on the right among the chairs"), we evaluate the alignment between the candidate's displacement and the reference direction $\vec{R}_{ref}$. The confusion score $S_{c}$ is defined by the cosine similarity:$$S_{c}(O_i) =  \frac{\vec{v}_{conf} \cdot \vec{R}_{ref}}{\|\vec{v}_{conf}\| \|\vec{R}_{ref}\|} $$This score effectively measures the target's relative standing within its group. By identifying the object that occupies the described relative position among its peers, the module resolves ambiguities that global spatial relations cannot address. Consistent with our previous strategy, we compute $S_{c}$ for all candidates to avoid error propagation from premature filtering.

\begin{table*}[t]
  \caption{Comparative results on on the ScanRefer dataset. We report the Accuracy@0.25 and Accuracy@0.5 for Unique, Multiple, and Overall categories.}
  \label{tab:performance}
  \centering
  \begin{tabular}{llc cc cc cc}
    \toprule
    \multirow{2}{*}{Venue} & \multirow{2}{*}{Method} & \multirow{2}{*}{Type} & \multicolumn{2}{c}{Unique} & \multicolumn{2}{c}{Multiple} & \multicolumn{2}{c}{Overall} \\
    \cmidrule(r){4-5} \cmidrule(lr){6-7} \cmidrule(l){8-9}
    & & & Acc@0.25 & Acc@0.5 & Acc@0.25 & Acc@0.5 & Acc@0.25 & Acc@0.5 \\
    \midrule
    CoRL'24 & Vlm-grounder \cite{xu2024vlm}  & zero-shot & 66.0 & 29.8 & 48.3 & 33.5 & 51.6 & 32.8 \\
    CVPR'24 & 3DLFVG \cite{liu2026view} & zero-shot & 65.80 & 51.27 & 22.03 & 16.94 & 30.53 & 23.61 \\
    CVPR'24 & ZSVG3D \cite{yuan2024visual} & zero-shot & 63.8 & 58.4 & 27.7 & 24.6 & 36.4 & 32.7 \\
    CVPR'25 & SeeGround \cite{li2025seeground} & zero-shot & 75.7 & 68.9 & 34.0 & 30.0 & 44.1 & 39.4 \\
    ACM MM'25 & SeqVLM \cite{lin2025seqvlm} & zero-shot & 77.3 & 72.7 & 47.8 & 41.3 & 55.6 & 49.6 \\
    RA-L '24 & ViewInfer3D \cite{geng2024viewinfer3d} & zero-shot & - & - & 56.25 & 42.83 & 60.31 & 47.20 \\
    NIPS'25 & SPAZER \cite{jin2025spazer} & zero-shot & 80.9 & 72.3 & 51.7 & 43.4 & 57.2 & 48.8 \\
    \midrule
    CVPR'25 & TSP3D \cite{guo2025text} & fully & - & - & - & - & 56.45 & 46.71 \\
    CVPR'24 & VPP-Net \cite{shi2024aware} & fully & 86.05 & 67.09 & 50.32 & 39.03 & 55.65 & 43.29 \\
    CVPR'24 & $MA^2TransVG$ \cite{xu2024multi} & fully & 86.3 & 74.1 & 53.8 & 41.4 & 57.9 & 45.7 \\
    TCSVT'25 & Pseudo-EV \cite{geng2025pseudo} & fully & - & - & 55.56 & 43.00 & 61.42 & 47.96 \\
    AAAI'25 & LIBA \cite{wang2025liba} & fully & 88.81 & 74.27 & 54.42 & 44.41 & 59.57 & 48.96 \\
    AAAI'25 & AugRefer \cite{wang2025augrefer} & fully & 86.21 & 70.75 & 49.96 & 39.06 & 55.68 & 44.03 \\
    IJCAI'25 & DDPA-3DVG \cite{gu2025ddpa} & fully & 86.8 & 70.2 & 49.8 & 38.4 & 55.3 & 43.3 \\
    \midrule
    \rowcolor[gray]{0.9} \textbf{} & \textbf{TDVR(Ours)} & zero-shot & \textbf{80.05} & \textbf{72.85} & \textbf{65.48} & \textbf{58.93} & \textbf{70.85} & \textbf{64.06} \\
    \bottomrule
  \end{tabular}
\end{table*}

\subsection{Feature-Matched Category and Appearance Evaluation}
To locate objects whose categories match the description, we further feed the category label of each object in the scene graph and the category text of the target object from the structured description into a pre-trained BERT \cite{devlin2019bert} model, obtaining the embedding of the category label $E_{cat}$
 and the embedding of the target category $E_{tar}$
. We then compute the similarity between these two vectors as the category matching score $S_{cat}$ between each category and the target category.
In addition, we input the 2D cropped image of each object in the scene graph, together with the target object category text and descriptive text from the structured description, into a pre-trained CLIP model \cite{radford2021learning}, yielding the image feature vector $E_{pic}$
 and the appearance text feature vector $E_{\text{app}}$. By calculating the similarity between these two vectors, we obtain the appearance matching score $S_{app}$ for each object.

\subsection{Target Grounding}
We combine the obtained $S_{v}$ , $S_{c}$ , $S_{cat}$, and $S_{app}$ to compute the overall evaluation score $S_{total}$ for each object.

$$S_{\text{total}} = S_{\text{cat}} \cdot (\alpha S_v + \beta S_c + \gamma S_{\text{app}})$$

$S_{cat}$ is a metric that measures how similar the object category is to the description, and acts as a coefficient multiplied by the weighted sum of $S_{v}$, $S_{c}$ ,and $S_{app}$.
We then sort the final evaluation scores $S_{total}$ of all objects, and select the object with the highest total score as the target object.

\section{Experiments}

\subsection{Experiments Settings}

\noindent \textbf{Datasets.}
We evaluate our method on the ScanRefer \cite{chen2020scanrefer} and Sr3D \cite{achlioptas2020referit3d} datasets. Built upon ScanNet\cite{dai2017scannet}, ScanRefer \cite{chen2020scanrefer} provides 51,583 free-form natural language descriptions for 11,046 objects. These descriptions emphasize appearance attributes and complex spatial relationships, requiring models to achieve sophisticated cross-modal alignment between 3D point clouds and language embeddings.
ReferIt3D \cite{achlioptas2020referit3d} is a large-scale dataset for 3D visual grounding. It features two subsets: Sr3D \cite{achlioptas2020referit3d}, using synthetic spatial templates, and Nr3D\cite{achlioptas2020referit3d}, containing natural language. Since the Nr3D\cite{achlioptas2020referit3d} dataset does not provide camera pose information, whereas that of Sr3D\cite{achlioptas2020referit3d} can be inferred through its construction code, We therefore conduct our experiments on the Sr3D \cite{achlioptas2020referit3d} dataset.

\noindent \textbf{Implementation Details.}
We utilize Mask3D \cite{schult2023mask3d} for initial instance segmentation and scene graph node generation. For the query disambiguation and structured reasoning modules, we employ GPT-4o (temp=0.7) and DeepSeek-V3 (temp=0.3), respectively, to generate enriched descriptions and perform CoT-based parsing.We adopt a locally deployed all-MiniLM-L6-v2 (BERT-based) \cite{devlin2019bert} to compute text embeddings.In the viewpoint-aware directional reasoning module, the rotation angle of the scene uses 10 degrees as one rotation unit.
For visual-linguistic matching, the CLIP ViT-B/32 \cite{radford2021learning} model is used to extract features from both 2D object crops and appearance descriptions.During the target grounding stage, the fusion weights are empirically set as $\alpha=5$, $\beta=3$, and $\gamma=1$. All experiments are conducted on a single NVIDIA RTX 4090 GPU.

\begin{figure*}[t]
  \centering
  \setlength{\tabcolsep}{4pt} 
  \begin{tabular}{p{0.20\textwidth}cccc}
    \toprule
    \textbf{Description} & \textbf{Ground Truth} & \textbf{SeeGround} & \textbf{SPAZER} & \textbf{Ours} \\
    \midrule

    \parbox{0.20\textwidth}{\small (a) The \textbf{chair} is against the wall. it is the first \textbf{chair} \textcolor{myred}{from the right}.}
    & \adjincludegraphics[width=0.12\textwidth, valign=c]{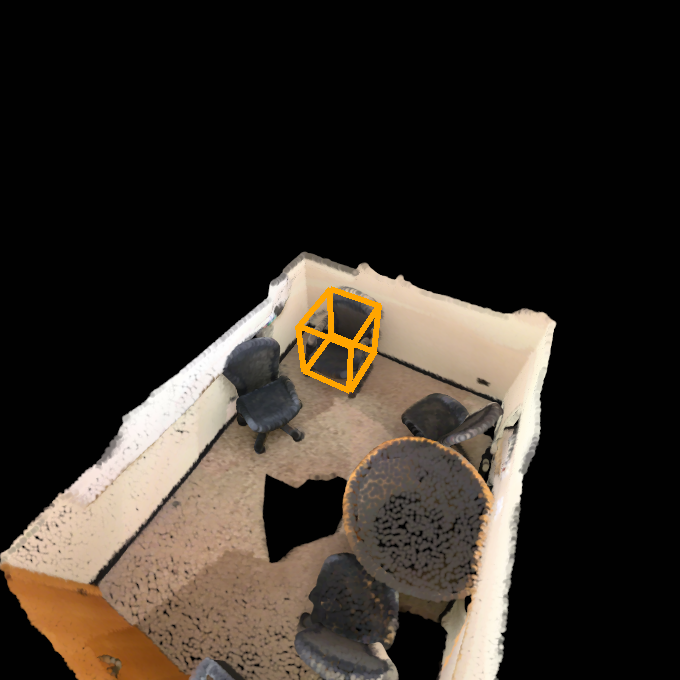}
    & \adjincludegraphics[width=0.12\textwidth, valign=c]{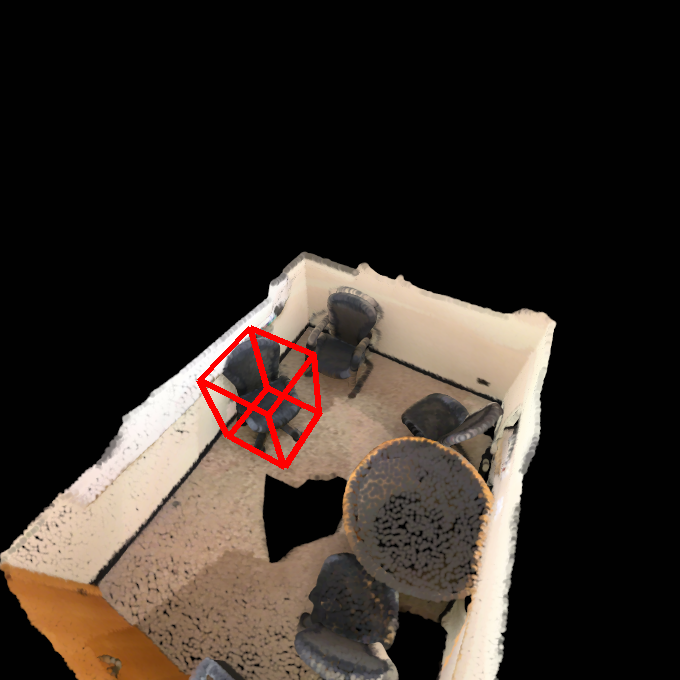}
    & \adjincludegraphics[width=0.12\textwidth, valign=c]{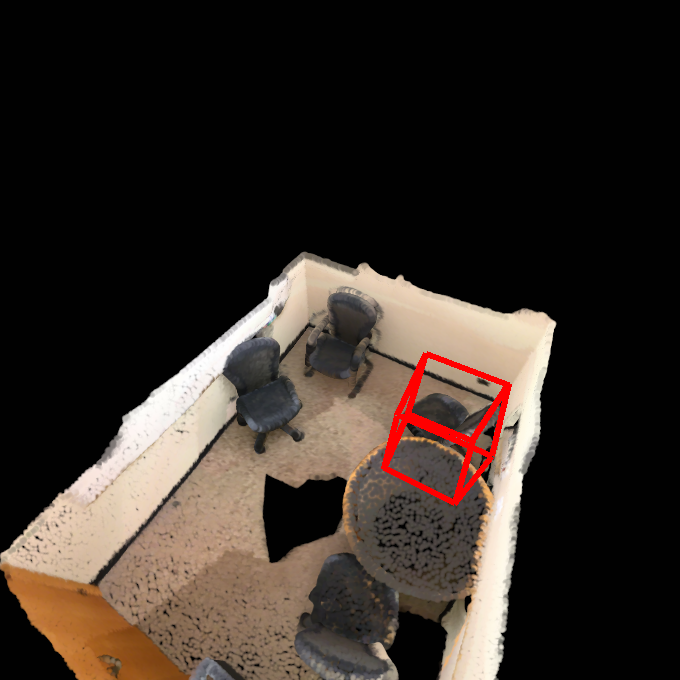}
    & \adjincludegraphics[width=0.12\textwidth, valign=c]{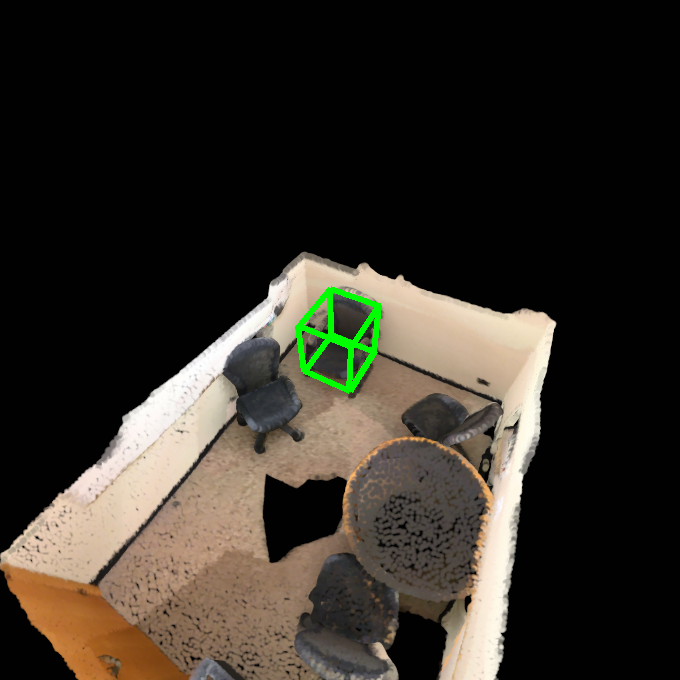} \\ \addlinespace[4pt] 

    \parbox{0.20\textwidth}{\small (b) this \textbf{chair} is facing away. \textcolor{mygreen}{it is wooden}.}
    & \adjincludegraphics[width=0.125\textwidth, valign=c]{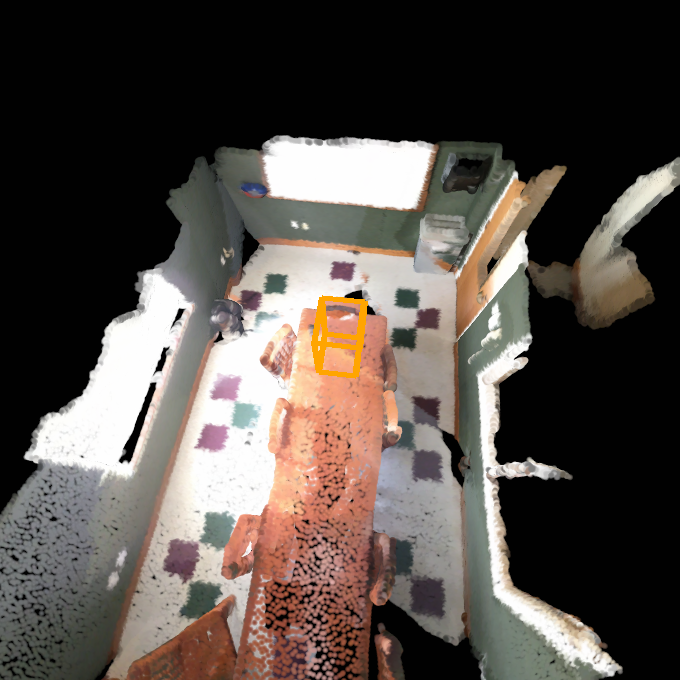}
    & \adjincludegraphics[width=0.125\textwidth, valign=c]{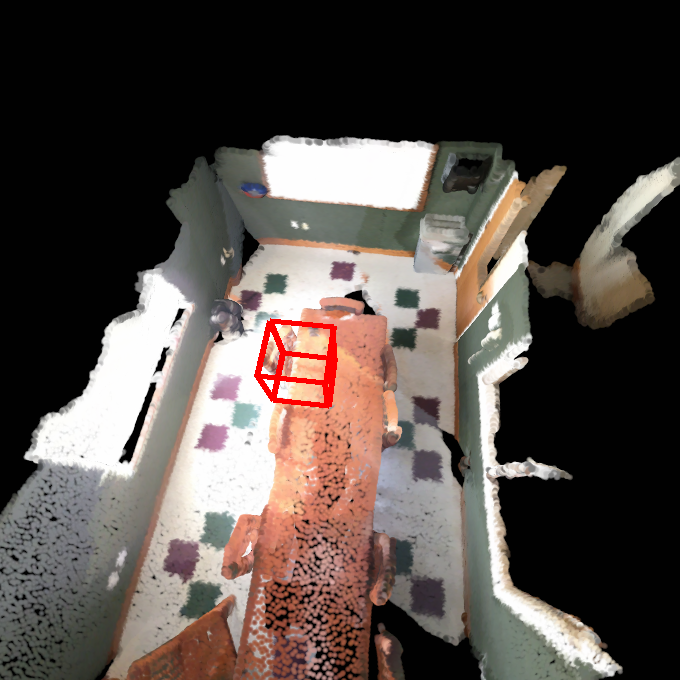}
    & \adjincludegraphics[width=0.125\textwidth, valign=c]{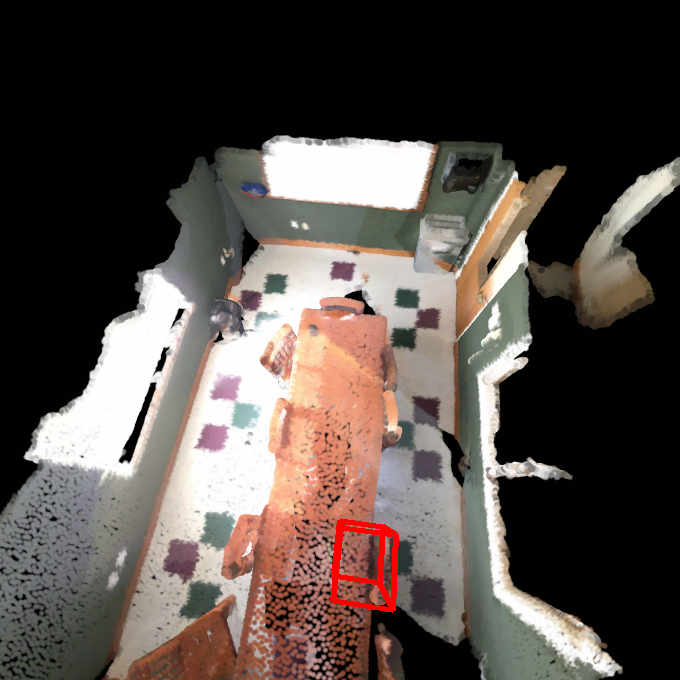}
    & \adjincludegraphics[width=0.125\textwidth, valign=c]{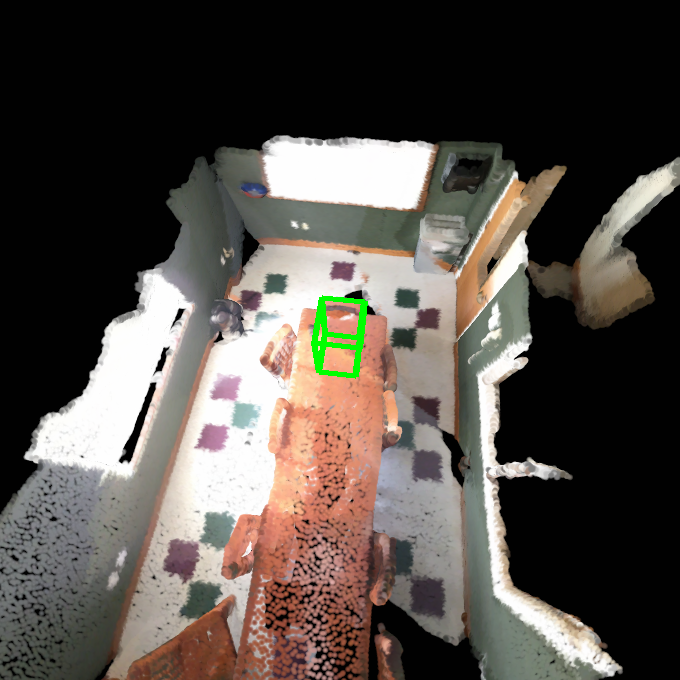} \\ \addlinespace[4pt]

    \parbox{0.20\textwidth}{\small (c) this is \textcolor{mygreen}{a blue} \textbf{sofa chair}. it is against a yellow wall.it is the first \textcolor{myred}{from the left}.}
    & \adjincludegraphics[width=0.12\textwidth, valign=c]{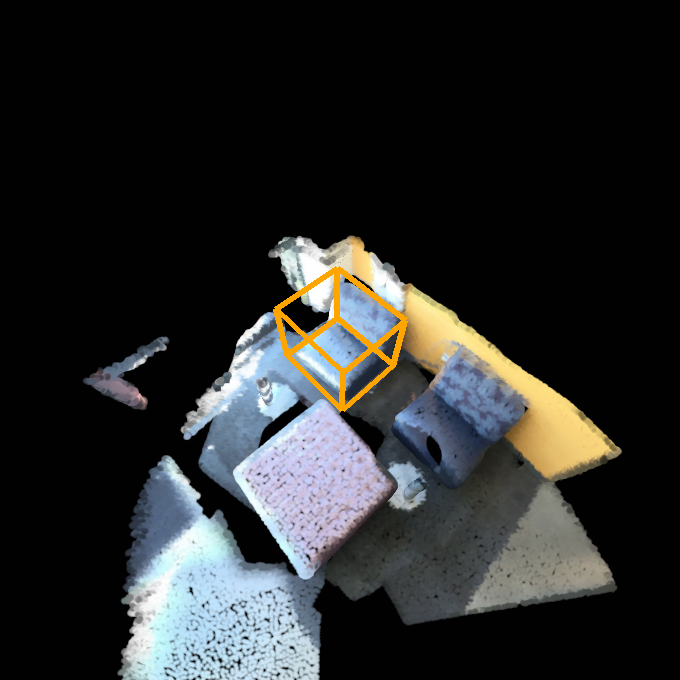}
    & \adjincludegraphics[width=0.125\textwidth, valign=c]{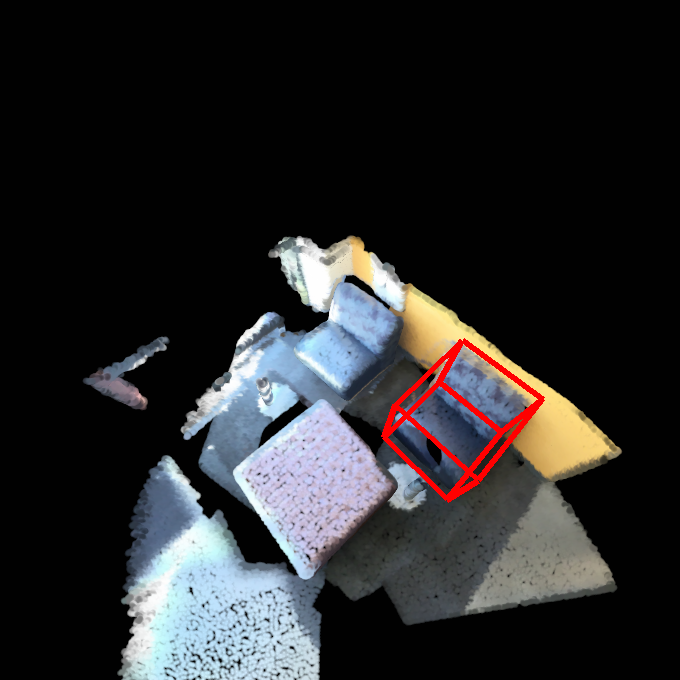}
    & \adjincludegraphics[width=0.125\textwidth, valign=c]{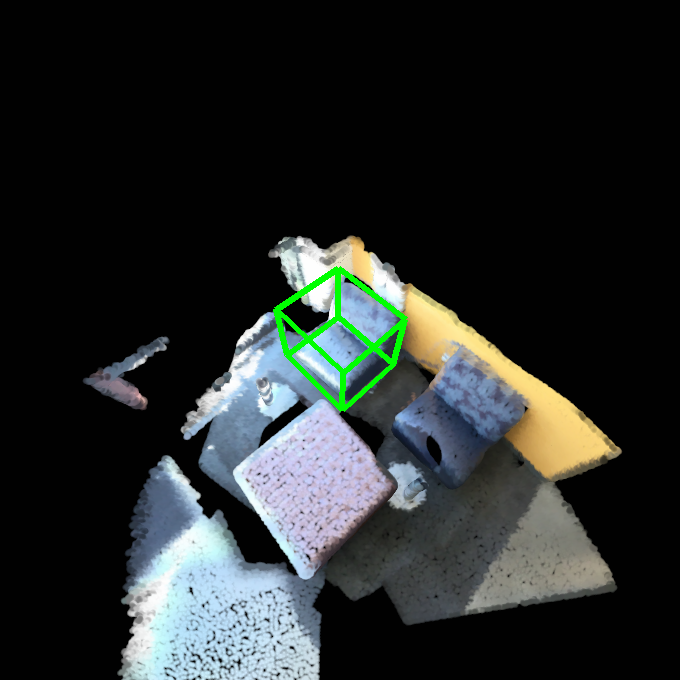}
    & \adjincludegraphics[width=0.12\textwidth, valign=c]{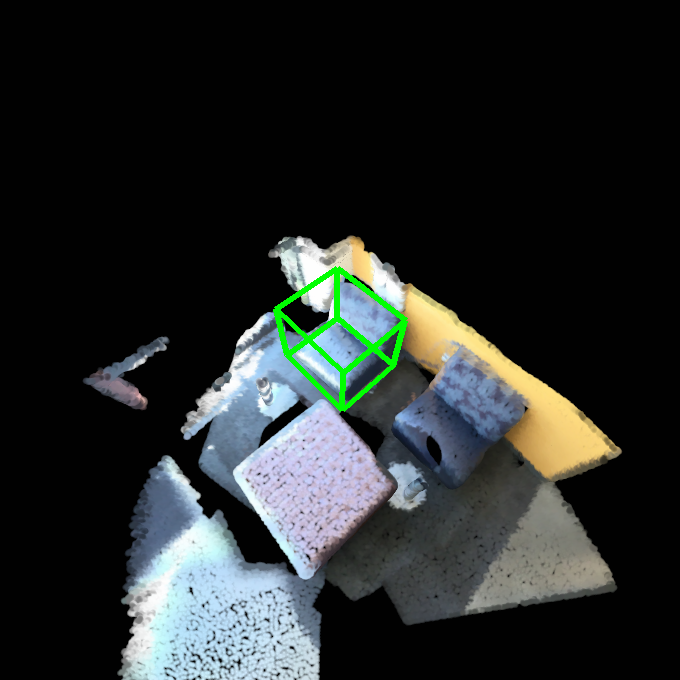} \\ \addlinespace[4pt]

    \parbox{0.20\textwidth}{\small (d) there is a \textcolor{mygreen}{rectangular} \textbf{pillow}. it is \textcolor{myred}{front of} a \textcolor{mygreen}{gray} \textbf{table} and \textcolor{myred}{left of} another \textbf{pillow}.} 
    & \adjincludegraphics[width=0.125\textwidth, valign=c]{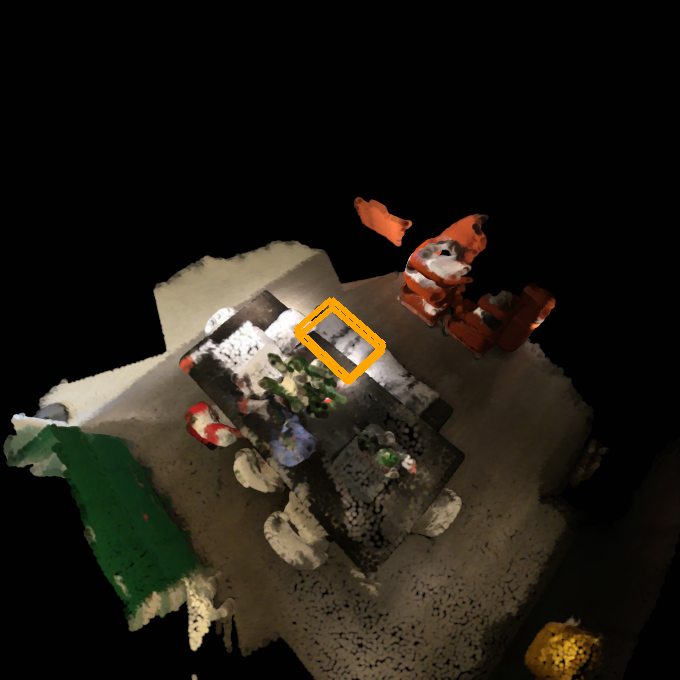}
    & \adjincludegraphics[width=0.12\textwidth, valign=c]{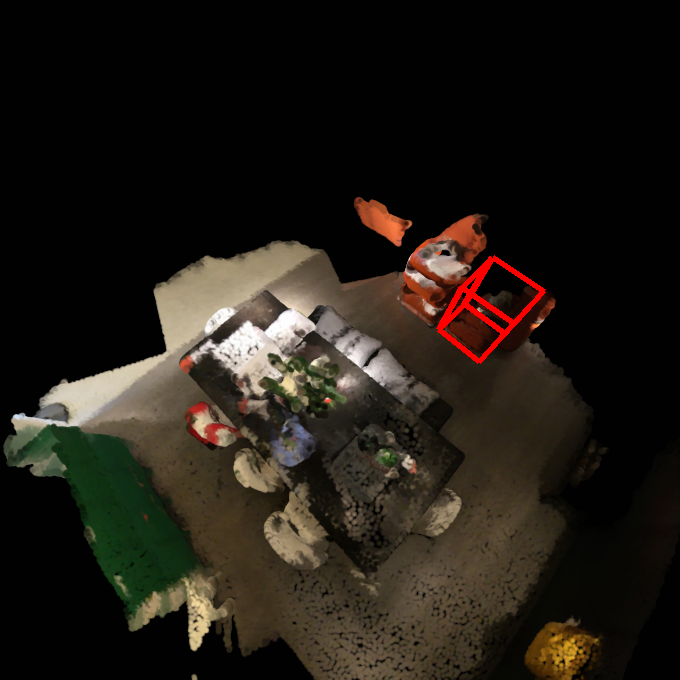}
    & \adjincludegraphics[width=0.125\textwidth, valign=c]{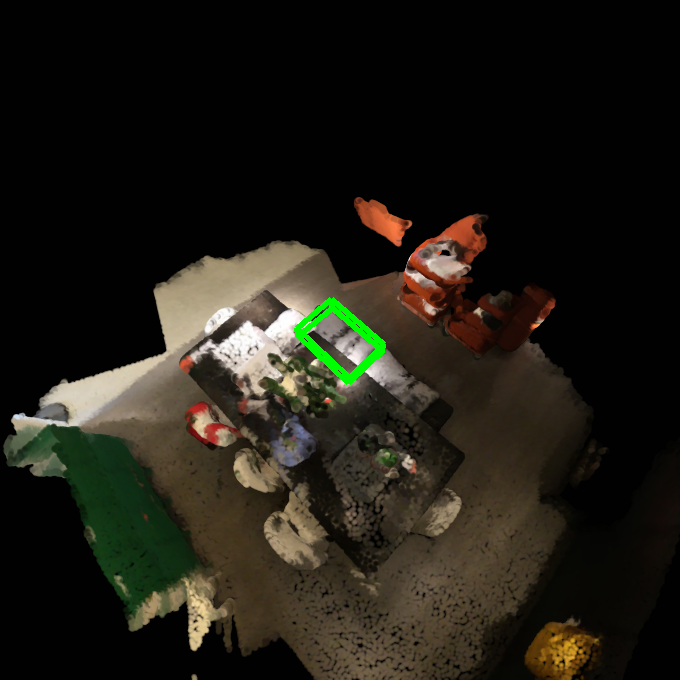}
    & \adjincludegraphics[width=0.125\textwidth, valign=c]{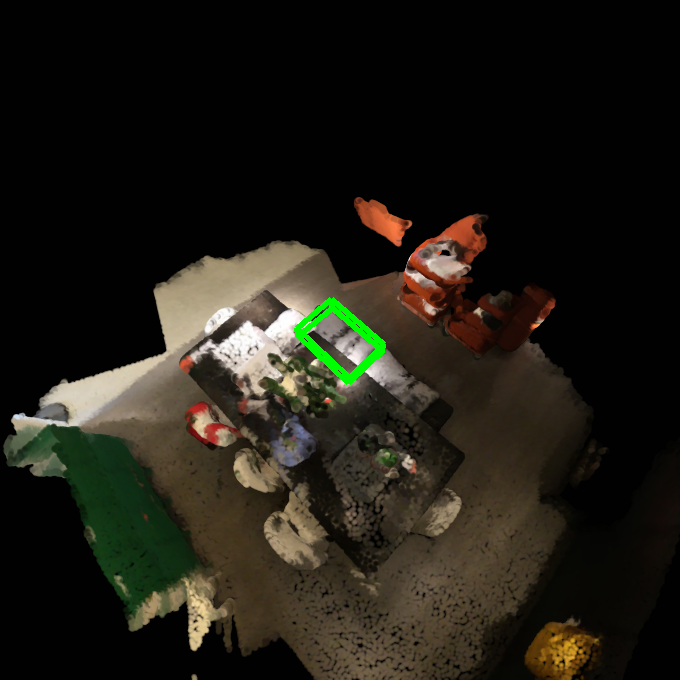} \\
    \bottomrule
  \end{tabular}
  \caption{Qualitative comparison of 3D visual grounding results.
  We visualize four challenging cases (a–d) that involve spatial relation descriptions and multiple similar confusing objects.Objects outlined in yellow boxes in the leftmost column represent the ground truth bounding boxes.
The three columns on the right show the results of SPAZER, SeeGround, and our proposed TDVR model, respectively.
Green boxes indicate correct predictions, while red boxes indicate incorrect predictions.
  }
  \label{fig:qualitative_results}
  \Description{A table-based visualization of comparative results from a qualitative evaluation of 3D object grounding methods, organized into four rows (cases) and five columns.

The table headers from left to right are: "Description", "Ground Truth", "SeeGround", "SPAZER", and "Ours".

Each row represents a specific grounding case, with a text description on the far left and corresponding visual results. The images in each column use a unique bounding box color to denote the detection result:

Ground Truth: Images with a gold or yellow bounding box, indicating the human-annotated correct object.

SeeGround: Images with a red bounding box, indicating the model's prediction.

SPAZER: Images with a bright green bounding box, indicating the model's prediction.

Ours: Images with a bright green bounding box (the same as SPAZER's), indicating the new model's prediction.

The rows are as follows:

Row (a)

Description: (a) The chair is against the wall. it is the first chair from the right.

Ground Truth: A gold box around the first chair from the right along the wall.

SeeGround: A red box around the table, incorrectly predicting the table as the chair.

SPAZER: A green box around the second chair from the right.

Ours: A green box around the first chair from the right, correct and identical to the ground truth.

Row (b)

Description: (b) this is a blue sofa chair. it is against a yellow wall.

Ground Truth: A gold box around the blue armchair against the yellow wall.

SeeGround: A red box around the table in front of the armchair.

SPAZER: A green box around the blue armchair, correct and identical to the ground truth.

Ours: A green box around the blue armchair, correct and identical to the ground truth.

Row (c)

Description: (c) this chair is facing away. it is wooden.

Ground Truth: A gold box around a specific wooden chair at the table.

SeeGround: A red box around a different, wrong wooden chair.

SPAZER: A green box around the correct wooden chair.

Ours: A green box around the correct wooden chair.

Row (d)

Description: (d) there is a rectangular pillow. it is front of a gray table and right of another pillow.

Ground Truth: A gold box around the specific correct pillow.

SeeGround: A red box around a different object in the background (perhaps another pillow or object).

SPAZER: A green box around the correct pillow.

Ours: A green box around the correct pillow.

The visualizations are presented against a black background. The "Ours" and "SPAZER" columns frequently match the Ground Truth, demonstrating good performance on these cases compared to SeeGround. In the first case (a), "Ours" is the only method to match the ground truth.}
\end{figure*}

\subsection{Quantitative Results}

As shown in Table \ref{tab:performance}, our method sets a new state-of-the-art performance for zero-shot 3D visual grounding on ScanRefer \cite{chen2020scanrefer} dataset. Under the overall metric, we achieve 70.85\% (Acc@0.25) and 64.06\% (Acc@0.5). Notably, our approach is the first zero-shot method to surpass the 60\% threshold on the strict Acc@0.5 metric, outperforming the previous best competitor, SPAZER \cite{jin2025spazer} (48.8\%), by a substantial margin of 15.26\%. Even when compared to recent fully-supervised methods (e.g., TSP3D \cite{guo2025text} and Pseudo-EV \cite{geng2025pseudo}), our zero-shot TDVR framework consistently yields superior performance, effectively narrowing the gap between zero-shot reasoning and supervised learning.
Our method demonstrates exceptional strength in the "Multiple" category, which contains distracting objects of the same class. We achieve 58.93\% (Acc@0.5), exceeding the previous best zero-shot method by 15.53\%. This significant improvement validates the effectiveness of our view-based similarity decoupled reasoning and viewpoint-aware directional reasoning modules in resolving spatial ambiguities and performing fine-grained discrimination among similar instances.
In the "Unique" category at Acc@0.25, our performance (80.05\%) aligns with top-tier methods but shows a marginal plateau. This is primarily due to the inherent recall limits of the frozen Mask3D \cite{schult2023mask3d} detector used for fair comparison across zero-shot baselines. However, under the stricter Acc@0.5 criterion, our method still leads with 72.85\%, proving that our reasoning pipeline provides higher-precision localization than existing supervised or heuristic-based alternatives.

\begin{table}[ht]
  \caption{Comparison with other methods on Sr3D dataset. We report the localization accuracy (\%) under different splits.\textsuperscript{$\dagger$} indicates results reproduced by our implementation.}
  \label{tab:results}
  \centering
  \setlength{\tabcolsep}{1.5pt} %
  \begin{tabular}{llccc}
    \toprule
    Method & Setting & Hard & View-dep. & Overall \\
    \midrule
    TSP3D (CVPR'25) \cite{guo2025text}& fully & - & - & 57.10 \\
    AugRefer (AAAI'25) \cite{wang2025augrefer} & fully & - & - & 60.22 \\
    DDPA-3DVG (IJCAI'25) \cite{gu2025ddpa} & fully & - & - & 69.70 \\
    \midrule
    SPAZER (NIPS'25) \cite{jin2025spazer}\textsuperscript{$\dagger$} & zero-shot & 49.00 & 55.65 & 57.52 \\
    SeeGround (CVPR'25) \cite{li2025seeground}\textsuperscript{$\dagger$} & zero-shot & 35.14 & 30.10 & 49.22 \\
    ZSVG3D (CVPR'24) \cite{yuan2024visual} \textsuperscript{$\dagger$} & zero-shot & 38.33 & 32.35 & 50.07 \\
    \textbf{Ours} & zero-shot & \textbf{63.23} & \textbf{79.03} & \textbf{70.00} \\
    \bottomrule
  \end{tabular}
\end{table}

We also conduct experiments on the Sr3D  \cite{achlioptas2020referit3d}dataset. Since the ReferIt3D \cite{achlioptas2020referit3d} dataset did not release camera pose information, the view annotations for the Sr3D \cite{achlioptas2020referit3d} dataset are inferred from its construction process.For objects with the $has\_{front}\_{direction}$ attribute set to True (i.e., oriented objects), we employ object orientations as viewpoint information. For objects with it set to False, the default direction of the point cloud is used as the viewpoint.
As shown in Table \ref{tab:results},among zero-shot methods of the same category, TDVR achieves a high accuracy of 70.00\%, outperforming many classic zero-shot approaches in recent years. Furthermore, we also surpass several representative fully-supervised methods, demonstrating the excellent grounding capability of our method. On the hard subset, TDVR reaches 63.23\%, which proves the superiority of our model in handling complex spatial relationships and confusing objects. And under the view-dependent subset, our performance achieves 79.03\%, reflecting the effectiveness of the model’s spatial perception ability.

\subsection{Qualitative Results}

To intuitively evaluate the localization capability of TDVR, we present qualitative comparisons with the representative methods (SPAZER \cite{jin2025spazer} and SeeGround \cite{li2025seeground}) in Fig. \ref{fig:qualitative_results}. The visualized scenes involve complex spatial reasoning and visually similar distractors.
In cases (a), (c),and (d), the descriptions involve viewpoint-dependent cues such as "first chair from the right". While SPAZER \cite{jin2025spazer} and SeeGround \cite{li2025seeground} struggle with the latent observation angle, our TDVR accurately identifies the target by inferring the observer's horizontal heading .
In scenes with multiple identical objects (e.g., the pillows in case (d) ,the chairs in case (b) and the sofas in case (c)), our method excels at resolving intra-category confusion.
In scenes where the original query is ambiguous, such as case (b), our method also achieves precise localization performance, demonstrating the excellent description disambiguation capability of the TDVR method.
TDVR maintains high-precision grounding where baseline methods often drift to adjacent distractors. As illustrated in Fig. \ref{fig:qualitative_results}, our method (green boxes) consistently aligns with the ground truth (yellow boxes), proving its superior robustness in ambiguous query reasoning and fine-grained 3D scene understanding.

\subsection{Ablation Studies}
\begin{table}[t]
  \centering
  \small
  \caption{Ablation study on the ScanRefer dataset. "OD": Observer-Centric Synergetic Disambiguation, "CM": Category Matching, "AM": Appearance Matching, "VR": viewpoint-Aware Directional Reasoning, "VSD":View-based Similarity Decoupled Reasoning
.}
  \label{tab:ablation}
  \setlength{\tabcolsep}{8pt} 
  \begin{tabular}{ccccccc} 
    \toprule
    ID & OD & CM & VR & VSD & AM & Overall \\
    \midrule
    1 & $\surd$ & $\surd$ & $\surd$ & $\surd$ & $\surd$ & \textbf{64.06} \\
    2 & $\surd$ & $\surd$ & $\surd$ & $\surd$ &  & 61.54 \\
    3 & $\surd$ & $\surd$ & $\surd$ &  &  & 59.21 \\
    4 & $\surd$ & $\surd$ &  &  &  & 39.95 \\
    5 & $\surd$ &  & $\surd$ & $\surd$ & $\surd$ & 59.71 \\
    6 &  & $\surd$ & $\surd$ & $\surd$ & $\surd$ & 40.59 \\
    \bottomrule
  \end{tabular}
\end{table}

We conduct a systematic ablation experiments to verify the contribution of each module in TDVR (refer to Table~\ref{tab:ablation}).
Removing the observer-centric synergetic disambiguation
module (ID 6 vs. 1) causes the most dramatic performance collapse, with overall Acc@0.5 dropping by 23.4\%. This underscores that resolving linguistic uncertainty is the prerequisite for precise 3D grounding.
The viewpoint-aware directional reasoning
(VR) module provides a substantial 19.3\% gain (ID 4 vs. 3), proving that viewpoint-aware spatial logic is the core driver for distinguishing similar objects. Additionally, the view-based similarity
decoupled reasoning
(VSD) (ID 3 vs. 2) further yields a 3.2\% boost, validating its necessity in fine-grained discrimination within the same category.

\begin{table}[h]
  \caption{Comparison of the Number of Anchors in Generated Descriptions in Observer-Centric Synergetic Disambiguation module on ScanRefer}
  \label{tab:anchors}
  \centering
  \small
  \setlength{\tabcolsep}{2pt} %
  \begin{tabular}{ccccccc}
    \toprule
    & \multicolumn{2}{c}{Unique} & \multicolumn{2}{c}{Multiple} & \multicolumn{2}{c}{Overall} \\
    \cmidrule(r){2-3} \cmidrule(lr){4-5} \cmidrule(l){6-7}
    number & Acc@0.25 & Acc@0.5 & Acc@0.25 & Acc@0.5 & Acc@0.25 & Acc@0.5 \\
    \midrule
    1 & 74.99 & 69.48 & 39.15 & 34.69 & 52.36 & 47.51 \\
    3 & 77.85 & 71.22 & 54.09 & 48.18 & 62.84 & 56.67 \\
    5 & 80.05 & 72.85 & 65.48 & 58.93 & 70.85 & 64.06 \\
    7 & 79.27 & 72.02 & 61.87 & 55.77 & 68.28 & 61.76 \\
    \bottomrule
  \end{tabular}
\end{table}

We investigate the sensitivity of spatial inference to the number of anchor objects ($N_{anc}$). As shown in Table \ref{tab:anchors}, performance peaks at $N_{anc}=5$. We observe a performance degradation when $N_{anc}$ increases to 7. This "more is less" phenomenon suggests that excessive anchors introduce redundant context, which potentially exceeds the optimal reasoning window of the LLM and introduces spatial noise that complicates the geometric matching process.

\begin{table}[h]
  \caption{Comparison of the Capabilities of Different Large Language Models on ScanRefer}
  \label{tab:comparison}
  \centering
  \begin{tabular}{lcc}
    \toprule
    LLM & Acc@0.25 & Acc@0.5 \\
    \midrule
    gpt-4o(SPAZE\cite{jin2025spazer})  & 57.20 & 48.80 \\
    \midrule
    gpt-4o & 59.31 & 52.71 \\
    glm-4 & 50.78 & 45.41 \\
    qwen2.5-72b-instruct & 47.58 & 42.43 \\
    \textbf{deepseekv3.2-reason (Ours)} & \textbf{70.85} & \textbf{64.06} \\
    \bottomrule
  \end{tabular}
\end{table}

We further evaluate TDVR across various LLMs to verify its generalizability (Table~\ref{tab:comparison}).DeepSeek-V3 \cite{liu2024deepseek} achieves the best performance, benefiting from its superior Chain-of-Thought reasoning in structured parsing.Notably, even when restricted to the same GPT-4o backbone, our TDVR framework outperforms the previous SOTA method, SPAZER \cite{jin2025spazer}, by 3.91\%. This result demonstrates that our performance gains stem from the inherent architectural innovations of the TDVR pipeline—specifically the multi-dimensional scene graph reasoning—rather than merely relying on a stronger language model.

\begin{table}[t]
  \centering
  \caption{Comparison of the influence of rotation unit angle on ScanRefer.}
  \label{tab:rotation_ablation}
  \small 
  \setlength{\tabcolsep}{3.5pt} 
  \begin{tabular}{lccccc}
    \toprule
    \textbf{Rotation Angle ($^\circ$)} & \textbf{10} & \textbf{20} & \textbf{30} & \textbf{40} & \textbf{50} \\
    \midrule
    Overall@0.5 (\%) & \textbf{64.06} & 60.82 & 58.64 & 57.33 & 46.61 \\
    \bottomrule
  \end{tabular}
\end{table}

We also conduct experiments on the angular unit of rotation, and perform viewpoint inference with different angles respectively.
The results are shown in Table \ref{tab:rotation_ablation}.
It can be observed that as the rotation unit angle increases, the accuracy of the final results decreases.
Therefore, we select $10^\circ$ as the rotation unit angle in this work.

\begin{table}[t]
  \centering
  \caption{Comparison of inference time among different methods.}
  \label{tab:time_comparison}
  \small
  \setlength{\tabcolsep}{16pt} 
  \begin{tabular}{lc}
    \toprule
    \textbf{Model} & \textbf{Time (s)} \\
    \midrule
    SPAZER & 23.5 \\
    Vlm-Grounder & 50.3 \\
    \textbf{Ours} & \textbf{10.21} \\
    \bottomrule
  \end{tabular}
\end{table}

As shown in Table \ref{tab:time_comparison} , we further compare the average inference speed of different models for a single query. Our model consumes less runtime than the representative models, taking only 10.21 seconds on average per query, which achieves substantial improvements over other zero-shot approaches.

\section{Conclusion}
We propose TDVR, a training-free framework for 3D visual grounding.
TDVR introduces an observer-centric synergetic disambiguation module, effectively alleviating language ambiguity.
To achieve accurate localization, TDVR constructs a multi-attribute scene graph and combines text-based structured reasoning to translate complex natural language into executable geometric constraints.
Furthermore, via the viewpoint-aware directional reasoning and view-based similarity decoupled reasoning modules, the system can infer the observer’s viewpoint and distinguish visually similar candidates in dense 3D scenes.
Extensive experiments on the ScanRefer and Sr3D benchmarks demonstrate that TDVR achieves state-of-the-art performance among training-free methods,significantly narrowing the performance gap between zero-shot and fully supervised 3D visual grounding approaches.

\begin{acks}

This work is jointly supported by the Fundamental Research Funds for the Central Universities (D5000250044, D5000250060),Natural Science Basic Research Program of Shaanxi (2025JC-YBQN-882,2025JC-YBQN-805),National Natural Science Foundation of China (62506301).

\end{acks}

\bibliographystyle{ACM-Reference-Format}
\balance
\bibliography{main}

\end{document}